%% file: root.tex
\documentclass[letterpaper, 10 pt, journal, twoside]{IEEEtran}  

\IEEEoverridecommandlockouts                              

\usepackage{graphics} 
\usepackage{epsfig} 
\usepackage{times} 
\usepackage{amsmath} 
\usepackage{amssymb}  

\usepackage{array}
\usepackage{booktabs}

\usepackage{hyperref}
\usepackage{caption}
\usepackage{xcolor,pifont}
\usepackage{color}
\newcommand{\newfinal}[1]{#1}
\newcommand*\colourcheck[1]{%
  \expandafter\newcommand\csname #1check\endcsname{\textcolor{#1}{\ding{52}}}%
}
\newcommand*\colourcross[1]{%
  \expandafter\newcommand\csname #1cross\endcsname{\textcolor{#1}{\ding{55}}}%
}
\colourcheck{green}
\colourcross{red}

\title{Bridging the Gap between Events and Frames through Unsupervised Domain Adaptation}

\author{Nico Messikommer, Daniel Gehrig, Mathias Gehrig, Davide Scaramuzza 
\thanks{Manuscript received: September, 9, 2021; Revised December, 7, 2021; Accepted January, 3, 2022.} 
\thanks{
This paper was recommended for publication by Editor Cesar Cadena upon evaluation of the Associate Editor and Reviewers' comments.
This work was supported by the National Centre of Competence in Research (NCCR) Robotics through the Swiss National Science Foundation (SNSF).
} 
\thanks{
The authors are with the Robotics and Perception Group, Department of Informatics, University of Zurich, and Department of Neuroinformatics, University of Zurich and ETH Zurich, Switzerland (\protect\url{http://rpg.ifi.uzh.ch}. {\tt\footnotesize nmessi@ifi.uzh.ch}). 
}
\thanks{Digital Object Identifier (DOI): see top of this page.}
}

\begin{document}

\maketitle

\input{sections/abstract}

\input{sections/introduction}
\input{sections/related_work}

\input{sections/method}

\input{sections/results}
\input{sections/conclusion}

\input{sections/supp_mat}

\bibliographystyle{IEEEtran}
{
\bibliography{IEEEabrv, all}
}

\end{document}

%% file: sections/abstract.tex
\begin{abstract}
Reliable perception during fast motion maneuvers or in high dynamic range environments is crucial for robotic systems.
Since event cameras are robust to these challenging conditions, they have great potential to increase the reliability of robot vision.
However, event-based vision has been held back by the shortage of labeled datasets due to the novelty of event cameras.
To overcome this drawback, we propose a task transfer method to train models directly with labeled images and unlabeled event data.
Compared to previous approaches, (i) our method transfers from single images to events instead of high frame rate videos, and (ii) does not rely on paired sensor data.
To achieve this, we leverage the generative event model to split event features into content and motion features.
This split enables efficient matching between latent spaces for events and images, which is crucial for successful task transfer.
Thus, our approach unlocks the vast amount of existing image datasets for the training of event-based neural networks.
Our task transfer method consistently outperforms methods targeting Unsupervised Domain Adaptation for object detection by 0.26 mAP (increase by 93\%) and classification by 2.7\% accuracy.
%

\end{abstract}

%% file: sections/introduction.tex
\vspace{-10pt}
\newfinal{
\section*{Multimedia Material}
The code of this project is available at  \url{https://github.com/uzh-rpg/rpg_ev-transfer}
Additional qualitative results can be viewed in this video: \url{https://youtu.be/fZnBSqni6PY} 
}

\vspace{-6pt}
\section{Introduction}
\IEEEPARstart{T}{he} outstanding properties such as high dynamic range, high temporal resolution, and low latency make event cameras promising for several robotic and automotive applications in edge-case scenarios, such as high dynamic range and fast relative motion. 
However, event cameras suffer from a recurring issue typical of any novel sensor modalities: the lack of labeled datasets.
Event-based datasets represent only 3.14\% of the existing vision dataset~\cite{Fischer21online, Yuhunang20eccv}.

\begin{figure}
\centering
\includegraphics[width=0.47\textwidth]{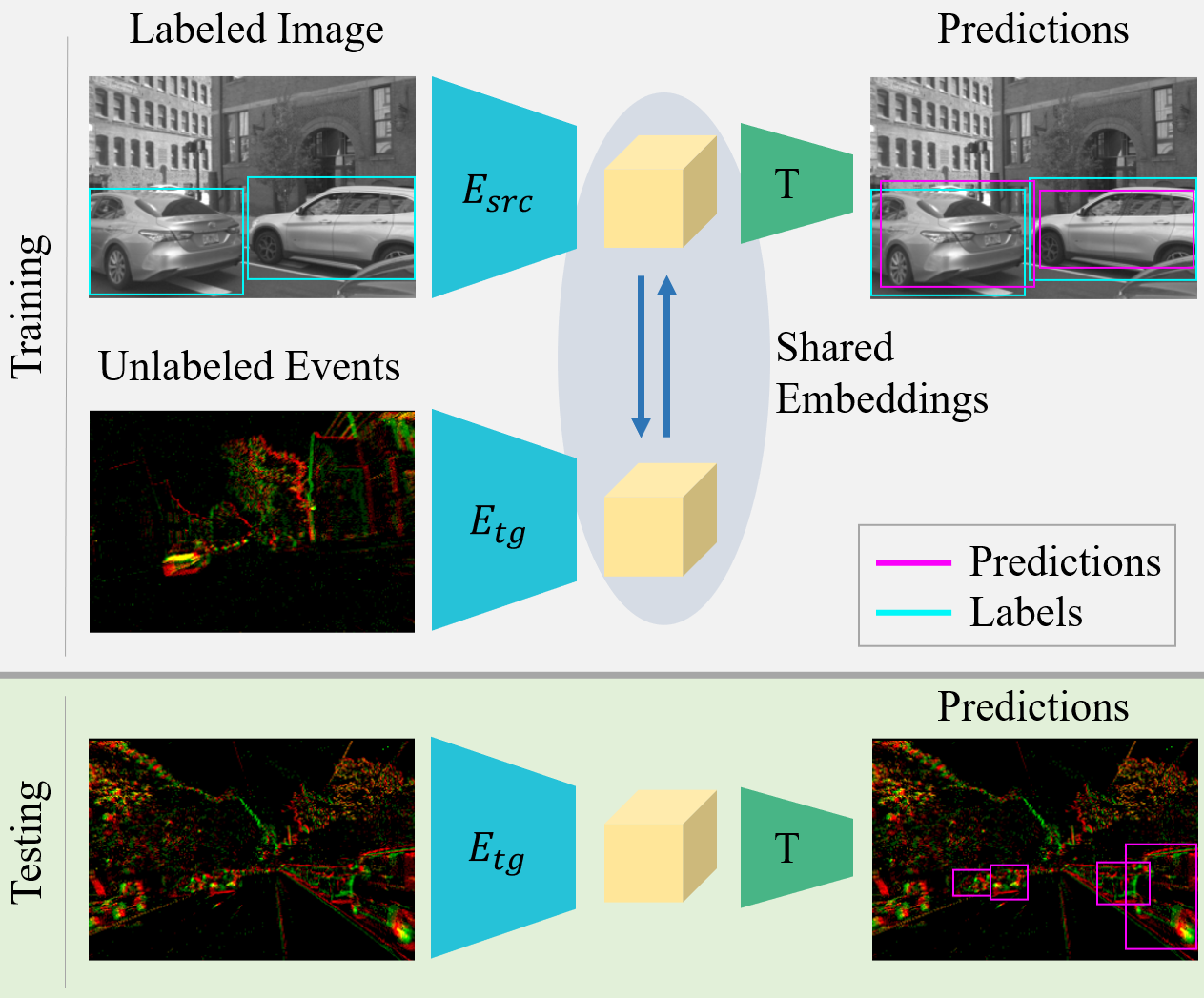}
\caption{
Our approach can teach a network to detect cars in event frames even though it was never told how cars look in the event domain.
This unsupervised domain adaption is possible by leveraging labeled grayscale images and unlabeled events.
During testing, our approach consists of a simple encoder $E$ and task network $T$ and thus has no computational overhead of first translating events to images.}\label{fig:eye_catcher}
\vspace{-15pt}
\end{figure}

Instead of capturing images at a fixed rate, event cameras measure changes of intensity asynchronously per pixel. This results in a stream of events that encodes the time, location, and polarity of the intensity change. For a more in-depth survey, we refer to~\cite{Gallego20pami}.
Despite the radical different working principle, the output of event and frame-based cameras still contains a significant information overlap, as both cameras share the underlying principle of capturing the scene irradiance through an optical system~\cite{Rebecq19pami}.

In this work, we show how this information overlap can be leveraged for \emph{Unsupervised Domain Adaptation (UDA)} of event-based networks, in which labeled source and unlabeled target data are available to transfer a task to the target domain. 
\newfinal{
Previous work has considered task transfer either between \emph{paired} events and frames, \emph{i.e.} recorded on the same pixel array~\cite{Yuhunang20eccv, Zhu19arxiv}, or through video-to-event translation~\cite{Gehrig20cvpr,Hu21cvprw}. These settings provide strong supervision between modalities, by providing per-pixel constraints or direct motion information from video, but severely limit the kinds of datasets which can be used. 
In fact, most large-scale datasets only comprise \emph{still images}~\cite{Russakovsky15ijcv, Ever10ijcv, sun2020scalability} instead of video, and were not recorded with a colocated event sensor, and are thus unpaired. Here, we present a method that can directly leverage these datasets, and does not rely on paired data or video inputs, opening up the immense bulk of frame-based datasets for event-based vision.
}

Furthermore, our approach does not rely on hand-designed generative models for video-to-events generation~\cite{Hu21cvprw, Gehrig20cvpr, Rebecq18corl} and has no computational overhead during inference by converting events to intensity images~\cite{Rebecq19pami}.

To bridge the gap between frames and events without paired data, we introduce a novel \emph{single-image-to-event} translation technique based on the event generation model combined with standard image-to-image translation techniques~\cite{Lee20ijcv, Zheng20eccv, Zhu17iccv}.
Crucially, instead of learning image-to-event translation directly, it only learns to correct initial guesses from the generative model.
However, event generation from a single image is ill-posed since motion information is missing. 
We solve this by explicitly extracting motion features from events in addition to the shared features that contain domain invariant information about the scene.
To achieve this split, we introduce a shared embedding discriminator and enforce shared feature consistency using sensor-specific knowledge.
Our approach can successfully transfer from images to events by leveraging large-scale datasets, which we show by outperforming state-of-the-art methods targeting UDA, for both object detection in MVSEC~\cite{Zhu18ral} and classification on N-Caltech101~\cite{Orchard15fns}.

Our contributions can be summarized as follows.
\begin{enumerate}
  \item We propose a transfer learning method that uses labeled frame-based datasets together with unlabeled events recorded in a target environment to train event-based networks. In particular, we show that networks trained on labeled daylight images can be transferred to challenging nighttime scenarios, where event cameras outperform standard cameras thanks to their higher dynamic range.
  \item Our approach leverages prior sensor knowledge based on the event generation model, and introduces a mapping from events to motion and content embeddings. This opens existing image-based datasets for event cameras, by transferring single images to events.
  \item Neural networks trained by our task transfer method outperform state-of-the-art object detection methods which target UDA by 0.27 mAP (an increase of 93\%) and achieve a 2.7\% accuracy increase in the classification task outperforming even some supervised approaches. 
\end{enumerate}


%% file: sections/related_work.tex
\section{Related Work}
\textbf{Unsupervised Domain Adaptation} 
%
The general problem of transfer learning based on labeled source data and unlabeled target data has accumulated vast literature over the years. For a survey, we refer the reader to~\cite{Wilson20tist}.
%
Early deep-learning-based methods use discriminators~\cite{Shen18aaai} or gradient reversal layers~\cite{Ganin15icml} to align the embedding space of source and target domain samples.
%

Analogously, \textit{image-to-image translation} methods can be used to induce shared embedding spaces by mapping samples from target to source domain, or vice versa. 
While the former enables the reuse of pre-trained networks trained in the source domain~\cite{Zheng20eccv}, the latter can provide a source of labeled datasets in the target domain by converting datasets from the source domain~\cite{LiPei18bmvc, Shrivastava17cvpr}. 

To directly transfer a task, recent work~\cite{Ryuhei20eccv,Hong18cvpr} proposed to jointly learn the task and mapping from input to domain-shared and domain-invariant features.
However, they can mostly work with shared network layers because they consider a smaller domain gap.
The domain gap between sensor modalities (RGB vs. Infrared) can also be exploited to supervise a RGB detection network in night sequences~\cite{Zanardi19icra}.

One work that addresses a large domain gap in the context of biomedical imaging is~\cite{Yang19miccai}, where domain-invariant features are constructed from source and target characteristics.

\textbf{Event-based Approaches}
%
To address the lack of labeled datasets, a recent class of methods seeks to convert events to high dynamic range (HDR) image reconstruction through supervised~\cite{Rebecq19pami}  or adversarial training~\cite{Mostafavi19cvpr,Mostafavi20cvpr,Wang20cvpr}.
With these images, standard pre-trained neural networks trained on images can be used. However, despite this advantage, these methods impose a computational overhead by first generating image reconstructions. 
Instead,~\cite{Yuhunang20eccv} simply adapt the first few layers of pre-trained frame-based networks to event data by enforcing feature consistency between the two separate sensor encoders. While this eliminates the need for costly event preprocessing, it requires paired images and events, \emph{i.e.} events and images recorded on the same sensor and scene to adapt a given network.
By contrast, the method in~\cite{Zhang20eccv} is designed to work with unpaired data but only converts between events in different illumination conditions.
The first works to leverage existing frame-based datasets were \textit{video-to-event translation} methods. 
These methods either rely on model-based~\cite{Rebecq18corl,Rebecq19pami,Gehrig20cvpr,Hu21cvprw} or data-driven~\cite{Zhu19arxiv} approaches to convert video sequences into artificial events, which can be used to directly train neural networks on event data.
This opened up the possibility of training networks for event data on larger and more diverse datasets. However, these methods are still limited to translating video to events, thus ignoring the majority of existing datasets that comprise images. 
The work most similar to ours leverages affinity graphs to perform the task transfer from frames to events~\cite{Wang21cvpr}.
In comparison, our approach splits the embedding space into shared and sensor-specific features and leverages the event generation model to align both domains.

In this work, we introduce a novel method that addresses the limitations of previous approaches by performing unsupervised domain adaptation, which \emph{(i)} maps unpaired images and events to a shared embedding space, \emph{(ii)} leverages single-image-to-event instead of video-to-event translation, and \emph{(iii)} performs task-transfer by jointly training a task-specific network on the shared embedding.
We introduce a novel single-image-to-event translation module that combines the event generation model~\cite{Gallego15arxiv} with standard translation methods.
Moreover, our method maps event data into separate content and sensor-specific features and only matches content features across modalities.
In doing so, we take inspiration from style-transfer techniques~\cite{Huang18eccv, Lee20ijcv, Zheng20eccv}.
%

%% file: sections/method.tex
\newcommand{\worlddomain}{\ensuremath{X}}
\newcommand{\worldstate}{\ensuremath{\mathbf{x}}}
\newcommand{\sensordomain}{\ensuremath{Y}}
\newcommand{\imgworlddomain}{\ensuremath{X_{\text{img}}}}
\newcommand{\eventworlddomain}{\ensuremath{X_{\text{event}}}}
\newcommand{\imgdomain}{\ensuremath{Y_{\text{img}}}}
\newcommand{\img}{\ensuremath{\mathbf{y}_{\text{img}}}}
\newcommand{\eventdomain}{\ensuremath{Y_{\text{event}}}}
\newcommand{\events}{\ensuremath{\mathbf{y}_{\text{event}}}} 
\newcommand{\latentf}{\ensuremath{\mathbf{z}}} 
\newcommand{\imglatentf}{\ensuremath{\mathbf{z}_{\text{img}}}}
\newcommand{\eventlatentf}{\ensuremath{\mathbf{z}_{\text{event}}}}
\newcommand{\eventspecific}{\ensuremath{\mathbf{\zeta}_{\text{event}}}} 
\newcommand{\taskbranch}{\ensuremath{T}}
\newcommand{\imgencoder}{\ensuremath{E_{\text{img}}}}
\newcommand{\eventencoder}{\ensuremath{E_{\text{event}}}}
\newcommand{\eventdecoder}{\ensuremath{D_{\text{event}}}}
\newcommand{\flowdecoder}{\ensuremath{D_{\text{flow}}}}
\newcommand{\rnet}{\ensuremath{R_{\text{ref}}}}
\newcommand{\latdiscr}{\ensuremath{F_{\text{lat}}}}
\newcommand{\eventdiscr}{\ensuremath{F_{\text{event}}}}

\section{Method}
\label{sec:method}

\input{latex_figures/method_overview}

Our goal is to train a network on labeled images for a specific task and transfer the network to events, such that the network successfully performs the task in the event-domain without requiring any labeled events nor paired images and event data, see Fig.~\ref{fig:eye_catcher}.
This setting of transferring a task from a labeled source domain (image domain $\imgdomain$) to an unlabeled target domain (event domain $\eventdomain$) is generally defined as Unsupervised Domain Adaptation, short UDA.
The task transfer is possible since event and frame-based cameras share the underlying principle of capturing the scene irradiance through an optical system.
Therefore, an information overlap exists on which the task can be learned on images and transferred to events.

In Sec.~\ref{sec:general_architecture}, we present the general network architecture and the latent space split into shared and sensor-specific features, which facilitates the task transfer.
The alignment of the shared latent space is enforced through multiple losses, which are explained in Sec.~\ref{sec:alignment_constraints}.
As a common constraint in the UDA literature~\cite{Chen19cvpr, Hoffman17icml, Murez18cvpr, Ryuhei20eccv, Yang19miccai, Zheng20eccv}, we perform domain translation to generate pseudo pairs, which are used to align the shared latent space on which the task is learned.
However, compared to classical UDA approaches, we only use a one-sided translation from images to events to achieve a better embedding alignment.
In Sec.~\ref{sec:img_to_event_translation}, we introduce our novel event construction from a single image based on the event generation model~\cite{Gallego15arxiv}, which is explained in Sec.~\ref{sec:prior_knowledge}.
Since the event generation model constitutes a relation between events, image gradients and optical flow, we can strongly constraint the image-to-event translation and thus significantly improve the task transfer. 
It is important to state that the image-to-event translation is only applied as an auxiliary task to help to transfer the task from images to events. 
As we directly optimize the task transfer from images to events, our method consistently outperforms pure translation methods~\cite{Gehrig20cvpr, Rebecq19cvpr, Rebecq19pami, Zhu19arxiv}.

\subsection{General Model Architecture}
\label{sec:general_architecture}
In our framework, events $\events$ and images $\img$ are processed with separate encoders $\imgencoder$ and $\eventencoder$ due to the large domain gap between $\imgdomain$ and $\eventdomain$, as shown in Fig.~\ref{fig:method_overview}. 
As the asynchronous output signal of event cameras also contains motion information, event cameras measure specific features $\eventspecific$ about the scene, which standard cameras can not perceive in a single frame.
This non-overlapping information, however, hinders the image-to-event task transfer as it is impossible to fully align the embedding space. 
We solve this by separating event features into \textit{sensor specific} features $\eventspecific$, which contain motion information, and \textit{content} features $\eventlatentf$, which carry information shared in both domains $\imgdomain$ and $\eventdomain$.
\begin{align}
\begin{split}
\label{eq:encoders}
    \imglatentf=&\imgencoder(\img) \\
    \eventlatentf=\eventencoder(\events) \quad&\quad \eventspecific=E_{\text{event, attr}}(\events).
\end{split}
\end{align}
The resulting shared features $\imglatentf$ and $\eventlatentf$ are given as input to the task branch $\taskbranch$, which computes the task-specific output. 
To generate pseudo event and image pairs, shared features from an image $\imglatentf$ are combined with event-specific features $\eventspecific$ from a random event sample to compute a pseudo-flow field using a flow decoder $\flowdecoder$.
The resulting pseudo-flow and the input image are then converted to events $\hat{\mathbf{y}}_{\text{event}}$ in the refinement network $\rnet$, 
\begin{equation}
\label{eq:decoder}
    \hat{\mathbf{y}}_{\text{event}}=\rnet(\flowdecoder(\imglatentf, \eventspecific)).
\end{equation}
Single-image-to-event translation is explained in more detail in Sec.~\ref{sec:img_to_event_translation}.
The overall architecture is depicted in Fig.~\ref{fig:method_overview}.

\subsection{Shared Latent Space Constraints}
\label{sec:alignment_constraints}
The unsupervised task transfer from images to events requires multiple constraints as there is neither task supervision in the event domain nor paired sensor data. 
Therefore, multiple losses are applied to enforce a shared latent space of $\imglatentf$ and $\eventlatentf$, which ensures that the task branch $\taskbranch$ successfully performs the task in both domains.
As a first constraint, we apply adversarial training~\cite{Goodfellow14nips} with a PatchGAN discriminator network $\latdiscr$~\cite{Isola17cvpr} to the latent features $\imglatentf$ and $\eventlatentf$
\begin{equation}
\begin{split}
\mathcal{L}_\text{lat.,disc.} =& \mathbb{E}_{\img}[\max(0,1-\latdiscr(\imglatentf))] \\
    &  + \mathbb{E}_{\events}[\max(0,1+\latdiscr(\eventlatentf))] \\
\mathcal{L}_\text{lat.,gen.}  =&  \mathbb{E}_{\imglatentf}[\latdiscr(\imglatentf)] -  \mathbb{E}_{\events}[\latdiscr(\eventlatentf)].
\end{split}
\end{equation}

Similar to~\cite{Zhu19arxiv}, we use a hinge-loss~\cite{Tran17nips} and optimize the above loss functions in an alternating fashion.
The above objective forces the latent space to be indistinguishable to the discriminator $\latdiscr$, and thus, the latent space becomes aligned.
As a consequence, the motion information required for the event generation can only be propagated through the event-specific features, which enforces the embedding split.

As an additional constraint, we generate pseudo sensor pairs using a one-sided translation from single images to events. 
These pseudo-pairs are used to formulate consistency losses on the latent variables $\imglatentf$ and $\eventspecific$, as summarized below: 
\begin{equation}
\begin{split}
    \mathcal{L}_{\text{cycle}} &= \vert \imglatentf - \eventencoder(\rnet(\flowdecoder(\imglatentf, \eventspecific))) \vert_1\\
    &+\vert \eventspecific - E_{\text{event, attr}}(\rnet(\flowdecoder(\imglatentf, \eventspecific))) \vert_1 \\
\end{split}
\end{equation}
To generate realistic events from a single image, the following adversarial loss is applied on the reconstructed events $\hat{\mathbf{y}}_{\text{event}}$ using an event discriminator $\eventdiscr$
\begin{equation}
\begin{split}
\mathcal{L}_\text{recons.,disc.} =& \mathbb{E}_{\hat{\mathbf{y}}_{\text{event}}}[\max(0,1-\eventdiscr(\hat{\mathbf{y}}_{\text{event}}))] \\
    &  \mathbb{E}_{\events}[\max(0,1+\eventdiscr(\events))] \\
\mathcal{L}_\text{recons.,gen.}  =&  \mathbb{E}_{\hat{\mathbf{y}}_{\text{event}}}[\eventdiscr(\hat{\mathbf{y}}_{\text{event}})]
\end{split}
\end{equation}

The used constraints are visualized with red arrows in Fig.~\ref{fig:method_overview}.
Overall, these general UDA methods represent a solid basis for closing the domain gap between events and images.
However, as shown in our experiments in Sec.~\ref{sec:experiments}, current UDA methods fall short in transferring task knowledge between the large gap of events and images. 
The next section shows how the event generation model can be leveraged to improve the task transfer between the sensor domains.

\subsection{Event Generation Model}
\label{sec:prior_knowledge}
The underlying principle of an event and frame camera can be exploited to guide the single-image-to-event translation.
As discussed in the introduction of Sec~\ref{sec:method}, event and frame cameras are both optical sensors, which capture the scene irradiance through lenses.
Due to this shared principle of measuring the light intensity, images can approximately be translated to events through the theoretical concept of the event generation model~\cite{Gallego15arxiv}.
This generative model describes the behavior of an ideal event camera under the assumption of constant brightness and of small time differences $\Delta t$.
In Eq. \ref{eq:event_eq}, $\Tilde{I}_{(x, y)}=\log(I_{(x, y)})$ expresses the measured intensity in log space and $\alpha$ represents the angle between the optical flow vector $\mathbf{{v}}_{(x, y)}$ and image gradient $\nabla \Tilde{I}_{(x, y)}$. 
\begin{equation} \label{eq:event_eq}
\begin{split}
\Delta\Tilde{I}_{(x, y)} & \approx - \langle \nabla \Tilde{I}_{(x, y)},  \mathbf{{v}}_{(x, y)} \Delta t \rangle \\
 & = - |\nabla \Tilde{I}_{(x, y)}| |\mathbf{{v}}_{(x, y)} | \Delta t \cos{\alpha} 
\end{split}
\end{equation}

An event is triggered if the log intensity change $\Delta\Tilde{I}$ is above a predefined contrast threshold $C$.
Thus, the number $N$ of events at a pixel $(x, y)$ can be approximated according to Eq.~\ref{eq:number_events}.
\begin{equation} \label{eq:number_events}
\begin{split}
N_{(x, y)} \approx \lfloor \Delta\Tilde{I}_{(x, y)} / C \rfloor
\end{split}
\end{equation}

The event generation model enables the transfer from a single image to events, if motion information in form of optical flow $\mathbf{{v}}_{(x, y)}$ and time difference $\Delta t$ is provided.
By considering only single frames, which most frame-based datasets consist of, the event generation is an ill-posed problem due to the missing motion information.
To account for that, we split the events $\events$ into two domains: latent space $\eventlatentf$ shared with image features and an additional sensor-specific space $\eventspecific$, in which the motion information in the events $\events$ is stored.
Thus, we can reconstruct artificial events from frames by combining content and sensor-specific features.

\vspace{-10pt}
\subsection{Event Generation based on Pseudo-Flow}
\label{sec:img_to_event_translation}
Following Eq.~\ref{eq:number_events}, we observe that the predicted events relate to the image gradient $\nabla \Tilde{I}_{(x, y)}$ via optical flow.
Instead of optical flow, we propose to directly predict pseudo-flow vectors $\hat{\mathbf{{v}}}_{(x, y)}$, which implicitly contain the unknown parameters $\Delta t$, $ \cos{\alpha}$ and $C$.
Thus, we do not need to compute these parameters explicitly.
\begin{equation} \label{eq:pseudo_flow}
\begin{split}
\hat{\mathbf{{v}}}_{(x, y)}  &= \mathbf{{v}}_{(x, y)}\frac{1}{C}\Delta t \cos{\alpha} \\
\hat{N}_{(x, y)} &\approx \langle \Delta\Tilde{I}_{(x, y)}, \hat{\mathbf{{v}}}_{(x, y)} \rangle
\end{split}
\end{equation}
It can be observed in Eq.~\ref{eq:pseudo_flow} that the number of predicted events $\hat{N}_{(x, y)}$ can either be changed by the pseudo-flow magnitude or by the angle between the two vectors $\Delta\Tilde{I}_{(x, y)}$ and $\hat{\mathbf{{v}}}_{(x, y)}$.
Thus, our pseudo-flow is not equivalent to optical flow as the adversarial training only enforces realistic events by either adjusting the direction or the magnitude of $\hat{\mathbf{{v}}}_{(x, y)}$.

The pseudo-flow is constructed from a combination of \emph{sensor-specific} and \emph{shared} features. 
The resulting pseudo-flow field adheres to the content extracted from an image $\imglatentf$ but with the general motion information of the event data, encoded in the \emph{sensor-specific} feature $\eventspecific$. 
Thus, the pseudo-flow generation also ensures that the \emph{sensor-specific} features are “flow-like” since they should produce realistic events, which is enforced by a discriminator.
The event generation based on pseudo-flow constrains the image-to-event translation and supports the adversarial training since it provides good event predictions early during training.

We propose a novel refinement block \rnet, which computes the inner product of the predicted pseudo-flow and image gradients as in Eq. \eqref{eq:pseudo_flow} to obtain an initial guess of the translated events.
The refinement net uses three convolutional layers to predict residual event representations, which are added to the initial guess. 
These residual events correct for overlapping polarity regions and event noise in the initial reconstruction.
The event decoder and refinement block are mainly supervised by the discriminator loss $\mathcal{L}_\text{recons.,disc.}$.
By enforcing realistic events, the supervision signal is back propagated to the event decoder since the fake events are generated based on the pseudo-flow and image gradient.

\subsection{Pseudo-Flow Augmentation}
\begin{figure*}
\centering
\begin{minipage}[b]{.7\textwidth}
\includegraphics[width=1.01\textwidth]{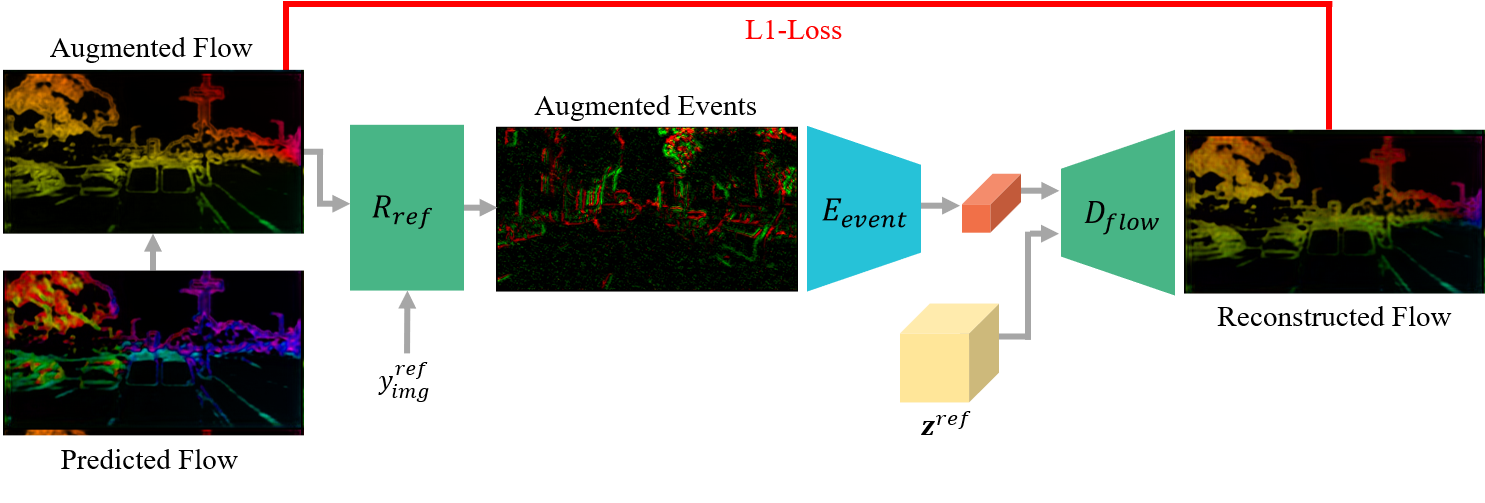}
\caption{
To enforce the split into sensor shared features and event-specific features, i.e., motion features, we propose to augment the pseudo-flow predictions. 
Specifically, we take a pseudo-flow prediction and augment the flow with the target domain-specific motions. In the shown car driving case, we sample an augmentation based on random epipoles in the image. Event-specific features ${\zeta}^{\text{aug}}_{\text{event}}$ are extracted from the events constructed based on the augmented flow.
These event specific features ${\zeta}^{\text{aug}}_{\text{event}}$ are then combined with the shared features ${z}^{\text{\text{ref}}}_{\text{\text{img}}}$ from the reference frame ${y}^{\text{ref}}_{\text{img}}$ to reconstruct the pseudo-flow.
An $L^{1}$-loss is then applied between the augmented and the reconstructed loss. 
By doing so, the networks can only adapt the motion features ${\zeta}^{\text{aug}}_{\text{event}}$ as the content features are fixed.
}\label{fig:flow_augmentation}
\end{minipage}\qquad
\begin{minipage}[b]{.26\textwidth}
\includegraphics[width=1.02\textwidth]{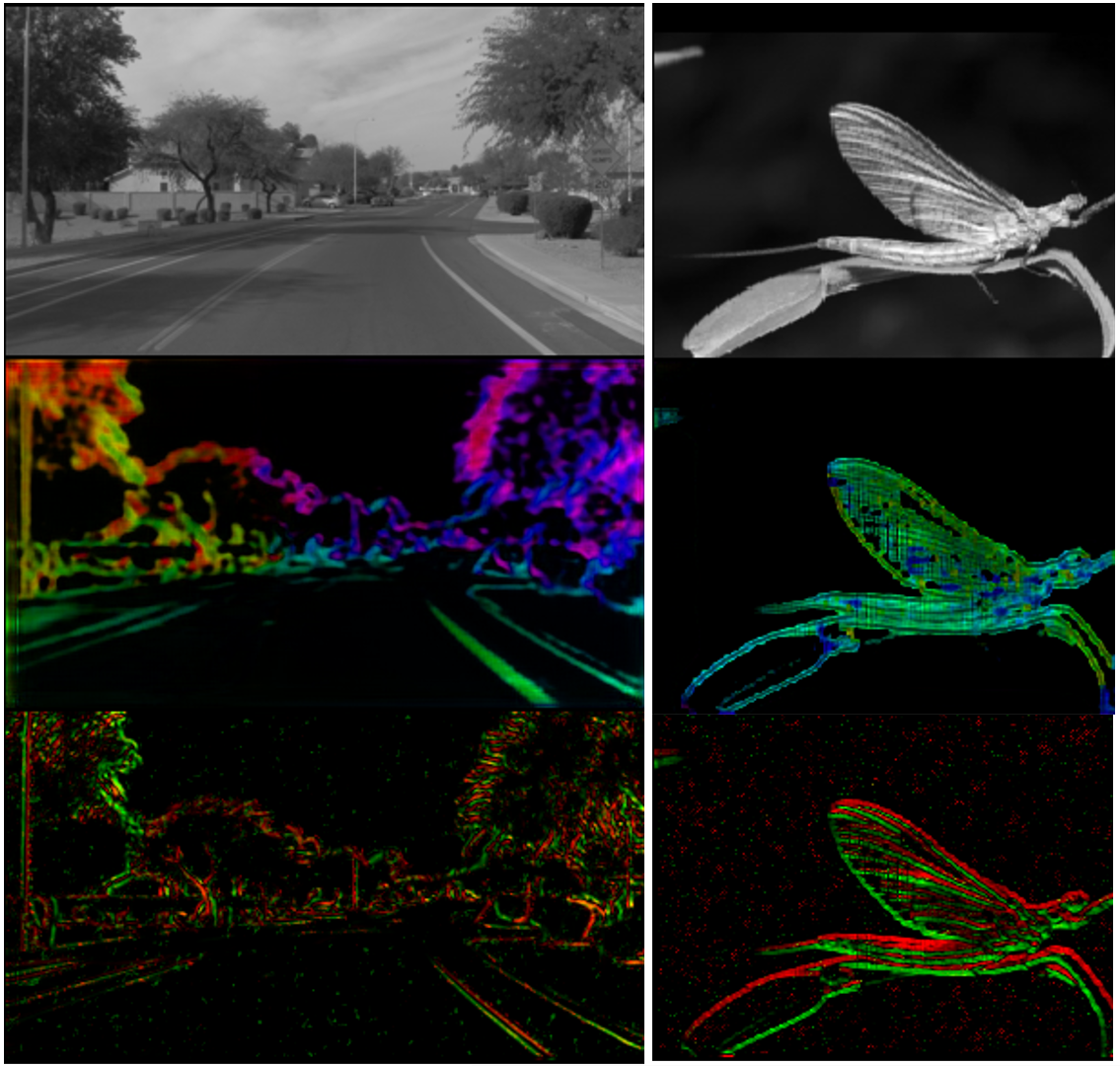}
\caption{
Images (top) are converted to events (bottom), via intermediate pseudo-flow map (middle) and image gradients.
}\label{fig:translation}
\end{minipage}
\end{figure*}

As the target domain is mostly known, we can augment the event generation by adding an artificial flow field according to the motion present in the target event data.
This augmentation loss $\mathcal{L}_\text{augm}$ is crucial to enforce the split into sensor-shared and \emph{sensor-specific} features.
It is essential to include augmentation consistent with the target event domain.
Otherwise, the discriminator easily distinguishes between the translated and real events, thus degenerating the adversarial training and the task transfer.
The augmented pseudo-flow field consists of vectors with a magnitude of the pseudo-flow predictions and the directions of the target-specific motion distributions. 
Events are then generated in the refinement module $\rnet$ from this augmented flow.
These events are then processed again by the event encoder $\eventencoder$ to obtain event-specific features $\eventspecific$, which are combined with the original shared feature to construct new events.
This event representation should be identical to the events obtained by the augmented flow as only the event-specific features $\eventspecific$ contain motion information.
Thus, an $L^{1}$-loss is applied between those two event representation, as visualized in Fig.~\ref{fig:flow_augmentation}.
The augmentation of reconstructed events can only be propagated through the event-specific features to reconstruct the augmented event representation since the content features stay constant, which further enforces the split into sensor-shared and sensor-specific features.
Moreover, the proposed flow augmentation can only be observed in the event representation, which means that the motion-specific attributes have to be extracted from an event sample. 
Thus, the augmentation loss would not be feasible if there is no unpaired event data.
The augmentation loss also helps preventing mode collapse of the event generation since it forces the event decoder to predict a large distribution of flow maps.

\subsection{Summary Loss Constraints}
\label{sec:final_loss}
In addition to the above introduced loss constraints, the task loss $\mathcal{L}_{task}$ is applied on the images and the fake events, which both have corresponding image labels. 
Eq.~\ref{final_loss} shows a summary of the presented loss functions.
\begin{equation} \label{final_loss}
\begin{split}
L_{Gen}  =&  \mathcal{L}_{\text{lat,gen.}} +\mathcal{L}_\text{recons.,gen.} +  \mathcal{L}_{\text{cycle}} + 2\mathcal{L}_{\text{augm}} + \mathcal{L}_{task}\ \\
L_{Dis} =& \mathcal{L}_{\text{lat,disc.}} + \mathcal{L}_\text{recons.,disc.} 
\end{split}
\end{equation}

%% file: latex_figures/method_overview.tex
\begin{figure*}
\centering
\includegraphics[width=0.97\textwidth]{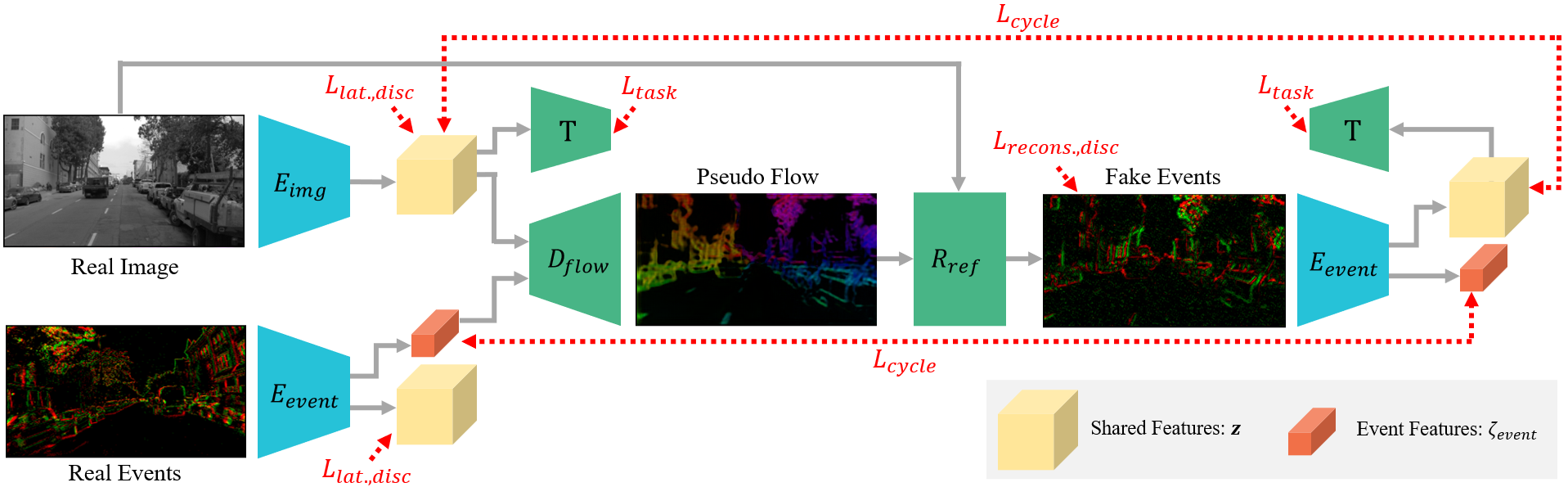}
\caption{
During training, our method uses single-image-to-event translation to transfer a task from the image to the event domain.
As there is a large domain gap between events and grayscale images, we use two separate encoders $\imgencoder$ and $\eventencoder$ (blue) to process unpaired images and event frames.
Shared features $\latentf$ (yellow) and event-specific features $\eventspecific$ (orange) are extracted from the event frame.
Both features are given as input to the event decoder $\flowdecoder$, which creates a pseudo flow map.
This flow map is combined with the image to create clean events using the event generation model.
To model sensor noise, the refinement module takes additionally random input feature maps.
The task network $T$ takes shared features from the images and the fake events as input and generates task predictions, which are supervised with the image labels.
The applied loss constraints are visualized with red arrows.
During inference, only the event encoder $\eventencoder$ and the task network $T$ are required, both are computationally light-weight networks.}

\label{fig:method_overview}
\vspace{-15pt}
\end{figure*}

%% file: sections/results.tex
\section{Experiments}
\label{sec:experiments}
\newcolumntype{C}[1]{>{\centering}m{#1}}

\subsection{Classification}
\label{sec:classification}
\textbf{Experimental Setup}
We validate our approach for event classification on the Neuromorphic-Caltech101 (N-Caltech101)~\cite{Orchard15fns} dataset.
Since Caltech101~\cite{Li06pami} samples were recorded for N-Caltech101, it is a straightforward choice to use Caltech101 as a labeled source dataset.
It is important to point out that we do not use paired sensor data even though it is available for N-Caltech101 and Caltech101.
As our approach can leverage unpaired single images for event-based training, we extend the frame-based Caltech101 with a set of additional images showing the 101 classes.
These additional images were also used in the baseline method VID2E~\cite{Gehrig20cvpr} for training a classification network on events, which are generated based on simulated motion.

Our task network for classification is inspired by Resnet18~\cite{He16cvpr}. 
In particular, we use the first layers up to the third residual block of Resnet18 without the first max-pooling layer for both sensor encoders.
The second of these residual blocks is shared between the event and image encoder.
The task network consists of the remaining Resnet18 layers.
All the Resnet18 layers were initialized with weights pre-trained on ImageNet~\cite{Russakovsky15ijcv}.
The architectures proposed in Drit++~\cite{Lee20ijcv} were adopted for the sensor-specific encoder, decoder, shared latent discriminator, and event frame discriminator.
While several modules are used during training, crucially, during testing, we only use the fast ResNet-18. which needs 4.5ms to process one sample on a Nvidia Quadro RTX 8000.
The event histogram~\cite{Maqueda18cvpr} is used as event representation to facilitate the image-to-event translation.
We augment the pseudo-flow with random translation fields (Section~\ref{sec:prior_knowledge}) as N-Caltech101 only contains translational motions.
%
%
%
Similar to supervised learning, we split the target data into training, validation, and test data.
Thus, no testing sample was seen during training/validation, neither in the image nor event domain.

\textbf{Results}
%
%
%
%
In addition to the state-of-the-art methods VID2E~\cite{Gehrig20cvpr} and E2VID~\cite{Rebecq19pami} applicable to the UDA setting, we also include supervised methods, which have access to the event labels during training.
The classification accuracies are reported in Tab.~\ref{tab:ncaltech101}.

Our approach outperforms the state-the-art method E2VID by 2.7\% in terms of accuracy.
Moreover, our inference network is a simple Resnet18, which is computationally much more lightweight than E2VID.
Compared to VID2E, our approach achieves a 4.1\% higher accuracy.
There are two possible reasons for the increased performance.
First, our method focuses specifically on task transfer and thus exploits the image and event domain to learn task-relevant features.
This multi-modal learning helps to extract more informative features, which was confirmed in recent work~\cite{Sayed18gcpr} as well.
Second, VID2E generates events based solely on model assumptions, whereas our approach uses the generative event model combined with a data-driven network to approximate target events.
Thus, our network can adjust better to the specific target event data, which is influenced by the event camera model and parameter settings.
As shown in Fig.~\ref{fig:translation}, our approach generates realistic event frames based on single grayscale images.

Our approach even outperforms the supervised methods HATS~\cite{Sironi18cvpr} and EST~\cite{Gehrig19iccv}, as seen in Table~\ref{tab:ncaltech101}.
One advantage of our method compared to HATS and EST is the increased size of the training dataset.
As our approach can use single images without corresponding events, we can easily extend the training dataset with additional class samples, as done for VID2E.
This also explains the higher classification accuracy of our approach compared to the supervised setting with the same architecture.
Since EvDistill~\cite{Wang21cvpr} was trained on a different training split for Caltech101 in the image domain, it is hard to compare against this approach.
Nevertheless, we report the performance of our approach trained on the complete Caltech101 dataset and the performance reported in~\cite{Wang21cvpr}.
We additionally report the performance of a simple cycle translation UDA framework without the embedding split into shared and event-specific features.
The significantly lower performance shows that the feature space split is crucial for the task transfer between images and events.
In the case of classification on N-Caltech101, the flow prediction does not provide any improvement compared to a model, which directly predicts an event representation.
This can be explained due to the simple planar motion distribution present in the event samples of N-Caltech101.
For a more complex motion distribution, i.e., in driving car sequences, the flow prediction almost doubles the object detection performance, see Tab.~\ref{tab:mvsec}.
%

\begin{table}
\centering
\begin{tabular}{m{2.7cm}C{1.2cm}>{\centering\arraybackslash}m{2cm}}

Method &  UDA & Accuracy $\uparrow$ \\
 \hline
E2VID~\cite{Rebecq19pami} & \greencheck & 0.821 \\
VID2E~\cite{Gehrig20cvpr} & \greencheck & 0.807 \\
Simple Cycle & \greencheck & 0.577\\
Ours w/o Flow & \greencheck & \textbf{0.848} \\
Ours & \greencheck & \textbf{0.848} \\
\hline
E2VID~\cite{Rebecq19pami} & \redcross & 0.866 \\
VID2E~\cite{Gehrig20cvpr} & \redcross & 0.906 \\
EST~\cite{Gehrig19iccv} & \redcross & 0.817 \\
HATS~\cite{Sironi18cvpr} & \redcross & 0.642 \\
Ours supervised & \redcross & 0.839 \\
\hline
EvDistill\text{*}~\cite{Wang21cvpr} & \greencheck & 0.902 \\
Ours\text{*} & \greencheck & 0.938 \\

\end{tabular}

\caption{\label{tab:ncaltech101} Classification accuracies on the N-Caltech101 dataset. The top of the table shows the methods targeting UDA, i.e., they do not have access to the event labels during training. We have also listed methods that use the ground truth labels during training and are thus not applicable for UDA. To stay consistent with the evaluation in \cite{Wang21cvpr}, we report the performance achieved by our model trained on the whole Caltech101 dataset(\text{*}).}
\end{table}

\begin{table}
\centering
\begin{tabular}{m{3.45cm}C{1.3cm}>{\centering\arraybackslash}m{1.8cm}}

Method & Unpaired & mAP $\uparrow$ \\
 \hline

ESIM~\cite{Rebecq18corl} & \greencheck & 0.02 \\
E2VID~\cite{Rebecq19pami} & \greencheck & 0.28 \\
Ours w/o flow & \greencheck & 0.26 \\
Ours w/o augm & \greencheck & 0.48 \\
Ours w/o split & \greencheck & 0.41 \\
Ours& \greencheck & \textbf{0.54} \\
\hline
EventGAN~\cite{Zhu19arxiv} & \redcross & 0.30 \\
YOLOv3-GN*~\cite{Yuhunang20eccv} & \redcross& 0.70 \\
\end{tabular}
\caption{\label{tab:mvsec} Mean average precision for the task of object detection on the MVSEC dataset. *Different test labels and trained on the same sequence}
\vspace{-15pt}
\end{table}
%
\subsection{Object Detection}
\label{sec:object_det}
\textbf{Experimental Setup}
In the case of object detection, we evaluate on the Multi-Vehicle Stereo Event Camera Dataset (MVSEC)~\cite{Zhu18ral}.
The authors of EventGAN~\cite{Zhu19arxiv} provided us with car bounding box labels for the outdoor\_day\_2 sequence, on which they evaluated EventGAN.
For training, we use the two outdoor sequences from MVSEC and add the DDD17~\cite{Binas17icml} dataset, which contains unlabeled events, both captured with a DAVIS346\cite{Brandli14ssc}.
%
As a labeled image dataset, we use Waymo Open Dataset~\cite{sun2020scalability}.
\newfinal{
In addition to MVSEC, we also report detection results on the large-scale, labeled 1 MP automotive detection dataset (1-MP-ADD)~\cite{Perot20nips}, which contains labels for 25 million object bounding boxes. Here, we train two methods, one using the labeled training data, and one in the UDA setting, namely, with unlabeled training data and labeled images from Waymo.
To stay consistent with other evaluation datasets, we only consider the "car" class in 1-MP-ADD.
}
Except for the task branch, we use the same network layers as for the classification task.
Similar to EventGAN and the network grafting approach YOLOv3-GN~\cite{Yuhunang20eccv}, the task branch for object detection consists of YOLOv3~\cite{Redmon18arxiv} layers.
During inference, our network needs 27ms to process one sample on a Nvidia Quadro RTX 8000.

\textbf{Results}
We compare against the event simulator ESIM~\cite{Rebecq18corl} as well as E2VID on the task of object detection.
Additionally, YOLOv3-GN and EventGAN are included as paired baselines, i.e., they were trained with events and the corresponding frames.
For the performance of YOLOv3-GN, we report the value published in their paper, which was computed on the same outdoor\_day\_2 sequence, but with different bounding boxes generated by a frame-based object detector applied to the grayscale images.
%
The object detection performances on MVSEC are reported in Table~\ref{tab:mvsec} as mean Average Precision (mAP)~\cite{Ever10ijcv}.
Compared to approaches trained on unpaired data, our approach achieves the highest performance, outperforming the next best method \cite{Rebecq19cvpr} by 26\% in terms of mAP. 
We credit the better performance to the fact that, while E2VID is only concerned with event-to-image translation, our method explicitly optimizes for the task objective, thus generating more task optimized representations.
The low performance of the detector trained using ESIM can be explained by the large domain gap between the generated and real events.
The events from ESIM were generated with a uniform planar motion~\cite{Zhu19arxiv}, which differs greatly from driving sequence motions.
%

%
%
We also compare our method against paired approaches~\cite{Zhu19arxiv,Yuhunang20eccv}, which use the grayscale images and the corresponding events of the outdoor\_day\_2 sequence during training.
Nevertheless, our method outperforms EventGAN by a significant margin (0.24 mAP). 
Compared to EventGAN, which generates labeled events from two frames, we focus on the specific task transfer from images to events. 
By splitting the embedding space into shared and event-specific motion features, we can leverage the labeled images to extract task-specific knowledge in the shared space.
Therefore, the task network can focus on task-specific features.
Moreover, the significant improvement can be attributed to the combination of our novel flow module (from 0.26 to 0.48), our flow augmentation (from 0.48 to 0.54), and our proposed feature split (from 0.41 to 0.54). 
As our method combines prior sensor knowledge with adversarial training, we can generate a more realistic event distribution and thus have an advantage at transferring the object detection task from images to events.
%
For YOLOv3GN, a pre-trained Yolov3 network was adapted directly to the frames from the outdoor\_day\_2 with the corresponding events to align the embedding space with paired data.
As the same sequence is used as a test set, a fair comparison is difficult since the reported test score is likely to overestimate the true test score.
By contrast, while our method has a 16\% lower performance, it is important to note that it does not have access to labels in the target domain and thus solely relies on unpaired images and labels from a vastly different domain. 
By contrast, the method in~\cite{Yuhunang20eccv}, achieves limited task transfer since it assumes paired (\emph{i.e.} per-pixel aligned and synchronized) data to work. 

\newfinal{
\textbf{Results on 1-MP-ADD}
The detector trained with our framework achieves 0.26 mAP compared to 0.48 mAP when trained in a supervised fashion on the 1-MP-ADD dataset.
Crucially, in this setting, unsupervised domain adaptation is not needed since labeled training data is available (25 million bounding boxes).  
However, if we evaluate the fully supervised network on MVSEC, it only achieves a performance of 0.49 mAP, which is lower than 0.54 mAP achieved by our UDA approach.
This highlights that our approach can more effectively bridge the domain gap between datasets compared to supervised methods.
}

\begin{figure}
\centering
\includegraphics[width=0.48\textwidth]{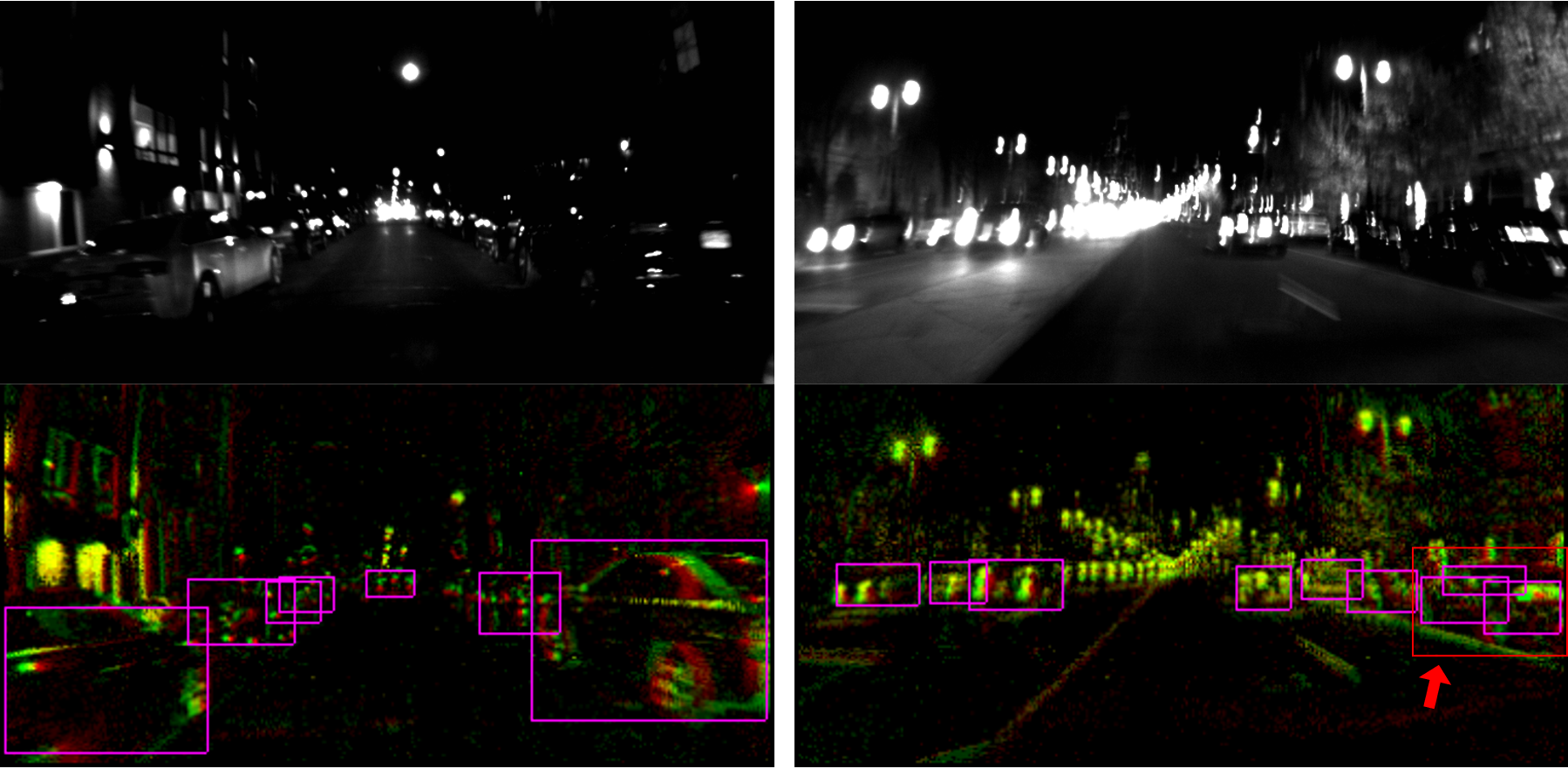}
\caption{Our task transfer framework enables to transfer from daylight images to events recorded during the night. 
The top row shows the VI-sensor images, which are underexposed (left) and suffer from motion blur (right). 
In contrast, the event histograms (bottom row) include much more details than the images, which helps the predictions of cars (magenta).
The bottom right arrow indicates the only three false predictions, which lay on top of just one car.}\label{fig:night_seq}
\end{figure}

\textbf{Daylight Images to Night-Time Events}
In general, events recorded during the night differ from day-light events~\cite{Hu21cvprw}.
Nevertheless, previous work~\cite{Perot20nips} qualitatively demonstrated that it is possible to transfer a network trained on daylight events to night events without any domain adaptation.
A fine-tuning on night events is usually also not possible since there are no labels available for night-time events~\cite{Perot20nips} due to the complicated labeling of night sequences.
Different from~\cite{Perot20nips}, our UDA framework explicitly leverages unlabeled events recorded during the night and creates a shared embedding space for daylight images and night-time events.
Thus, the task network trained with our framework can adapt to the noise distribution specific to nighttime events.
To demonstrate this ability, we use the relatively high-quality images from the Waymo Open Dataset to transfer the task of car detection to events recorded during the night, where standard cameras are underexposed.
We visualize our detector in this scenario on MVSEC~\cite{Zhu18rss} (Fig. \ref{fig:night_seq}).
Standard frames recorded with a VI-Sensor~\cite{Nikolic14icra}  (top row) are underexposed and blurry, while event data has a higher dynamic range and does not suffer from motion blur. 
Our method detects all cars present in the event stream (orange boxes), only making a single mistake by misidentifying a cluster of cars (red arrow).
Crucially, this method was entirely trained with labeled images in daylight scenarios, without ever seeing a label in the dark.
This example highlights the potential of the method for transferring task knowledge to challenging night-time scenarios. 
The robustness of our approach is also shown in the supplementary video.

\newfinal{
\subsection{Future Work}
We have shown the effectiveness of our proposed method on three event datasets, including translating and driving motions.
To extend our framework to datasets with different flow distributions, multiple potential flow augmentations could be applied such as flipping, rotating, scaling, and constant addition.
However, object motion independent of the global scene motion still remains a challenge for our method and is left as future work.
Additionally, the current method only uses the event-specific features during image-to-event translation. 
Their use in the task network remains unexplored, and could be useful for motion-related tasks such as optical flow estimation.
}

%% file: sections/conclusion.tex
\section{Conclusion}

Learning for event-based vision has been held back by the scarcity of training data. In contrast, image or video-based methods have tremendously improved in performance due to the availability of large-scale datasets. 
This work proposes a framework to address this problem by leveraging large-scale image datasets with unsupervised domain adaptation. To achieve this, our method transfers task-specific knowledge from frame-based datasets to the event-based domain without the need for paired sensor data. Therefore, this framework allows models to be trained directly with labeled images and \emph{unlabeled} event data. 
In conclusion, our work unlocks the potential to use any frame-based dataset to train an event-based network. 
By large-scale datasets like the extended N-Caltech101 dataset~\cite{Gehrig20cvpr} or the Waymo Open Dataset~\cite{sun2020scalability}, we outperform state-of-the-art method for classification by 2.7\% and object detection by 0.26 mAP, in the UDA setting.

%% file: sections/supp_mat.tex

\clearpage
\section*{\Large \bf Supplementary: Bridging the Gap between Events and Frames through \\Unsupervised Domain Adaptation}

\section*{A1 Losses}
In the following, we give a detailed explanation of the losses applied in our proposed framework. 

\subsection*{A1.1 Adversarial Losses}
\textbf{Latent Space Loss}

To ensure that the task branch $\taskbranch$ can seamlessly transfer between events and frames, we enforce its input, i.e., both latent representation $\imglatentf$ and $\eventlatentf$, to lay on one manifold.
This is done by applying adversarial training \cite{Goodfellow14nips} with a discriminator network \latdiscr.
The adversarial training aligns the distribution of $\imglatentf=\imgencoder(\img)$ and $\eventlatentf=\eventencoder(\events)$. 
Similar to \cite{Zhu19arxiv}, a hinge adversarial loss~\cite{Tran17nips} is adopted.
\begin{equation}
\begin{split}
\mathcal{L}_\text{lat.,disc.} =& \mathbb{E}_{\img}[\max(0,1-\latdiscr(\imglatentf))] \\
    &  + \mathbb{E}_{\events}[\max(0,1+\latdiscr(\eventlatentf))] \\
\mathcal{L}_\text{lat.,gen.}  =&  \mathbb{E}_{\imglatentf}[\latdiscr(\imglatentf)] -  \mathbb{E}_{\events}[\latdiscr(\eventlatentf)].
\end{split}
\end{equation}

\textbf{Image-to-event Translation Loss}

As an additional constraint, we force the latent representations $\eventspecific,\imglatentf$ to carry sufficient information, such that they can be decoded into an artificial event-frame. 
To do this, we combine the event-specific features $\eventspecific$ from a random event representation $\events$ and the content features $\imglatentf$ from an image $\img$ to generate artificial events $\hat{\mathbf{y}}_{\text{event}}=\rnet(\flowdecoder(\imglatentf, \eventspecific))$ using a pseudo-flow decoder \flowdecoder\ and a refinement net \rnet.
These events share the image content but contain the event-specific features, e.g., motion distribution, of the reference events $\events$. 
The following adversarial loss is applied on the image-to-event reconstruction $\hat{\mathbf{y}}_{\text{event}}$.
Similar to the embedding space alignment, a PatchGAN~\cite{Isola17cvpr} discriminator and the hinge loss~\cite{Tran17nips} is adopted for the sensor translation.

\begin{equation}
\begin{split}
\mathcal{L}_\text{recons.,disc.} =& \mathbb{E}_{\hat{\mathbf{y}}_{\text{event}}}[\max(0,1-\eventdiscr(\hat{\mathbf{y}}_{\text{event}}))] \\
    &  \mathbb{E}_{\events}[\max(0,1+\eventdiscr(\events))] \\
\mathcal{L}_\text{recons.,gen.}  =&  \mathbb{E}_{\hat{\mathbf{y}}_{\text{event}}}[\eventdiscr(\hat{\mathbf{y}}_{\text{event}})]
\end{split}
\end{equation}

\subsection*{A1.2 Translation Consistency}
By translating from single images to events, we can formulate consistency losses on the latent variables $\imglatentf$ and $\eventspecific$ as summarized below:
\begin{equation}
\begin{split}
    \mathcal{L}_{\text{cycle}} &= \vert \imglatentf - \eventencoder(\rnet(\flowdecoder(\imglatentf, \eventspecific))) \vert_1\\
    &+\vert \eventspecific - E_{\text{event, attr}}(\rnet(\flowdecoder(\imglatentf, \eventspecific))) \vert_1 \\
\end{split}
\end{equation}

\subsection*{A1.3 Flow Augmentation Loss}
As shown in Figure 4 in the manuscript, we augment the flow by combining the augmented, event-specific features ${\zeta}^{\text{aug}}_{\text{event}}$ with the shared features ${z}^{\text{\text{ref}}}_{\text{\text{img}}}$ from the reference frame ${y}^{\text{ref}}_{\text{img}}$. The resulting event representation should be identical to the events generated by the augmented flow.
By enforcing this identity, the network can only store the augmented flow information in the event-specific features ${\zeta}^{\text{aug}}_{\text{event}}$ since the shared features are constant.

\begin{equation}
\begin{split}
    \mathcal{L}_{\text{augm}} &= \vert E_{\text{event, attr}}({y}^{\text{aug}}_{\text{event}}) \\
                             &- E_{\text{event, attr}}(\rnet(\flowdecoder({z}^{\text{\text{ref}}}_{\text{\text{img}}}, {\zeta}^{\text{aug}}_{\text{event}}))) \vert_1
\end{split}
\end{equation}

\subsection*{A1.4 Flow Smoothness}
As common in the literature, we add a smoothness loss $\mathcal{L}_{\text{smooth}}$ to further constrain the pseudo-flow prediction. The smoothness loss consists of the Charbonnier loss function~\cite{Sun14ijcv} applied to the difference of a flow vector with its eight neighboring flow vectors (including diagonal neighbors):
\begin{equation}
\mathcal{L}_{\text{smooth}}=\sum_\mathbf{x} \sum_{ \mathbf{y}\in\mathcal{N}(\mathbf{x})}\rho(\mathbf{v}_{\mathbf{x}}-\mathbf{v}_{\mathbf{y}})
\end{equation}
Where $\mathbf{x}=(x,y)$ are image pixels, $\mathcal{N}(\mathbf{x})$ denotes the 8 neighbors of $\mathbf{x}$ and $\rho(x)=(\epsilon^\alpha + x^\alpha)^\frac{1}{\alpha}$. We use $\alpha=0.45$ and $\epsilon=0.001$. 
\subsection*{A1.5 Image Gradient Loss}
To prevent mode collapse of the image-to-event generation, a gradient loss $\mathcal{L}_{\text{grad}}$ is applied on the event reconstructions.
This loss penalizes areas that have few events and high image gradient as events are generated mainly by image gradients. 
We enforce this constraint by introducing the following loss:
\begin{equation}
    \mathcal{L}_{\text{grad}} = \sum_{(x,y)}\max(0, 0.7 - N_{(x, y)}) \vert\nabla I_{(x, y)}\vert
\end{equation}
Where we sum over all pixels that have an image gradient magnitude $\vert\nabla I_{(x, y)}\vert>0.7$.
Here $N_{(x,y)}\in[0,1]$ represents the normalized number of events per pixel, as predicted by the event generation module. Without this loss, the event generation focuses too strongly on the noisy and low-quality events present in the real event domain. 
This noise prediction helps the generator fool the discriminator since it can just predict noisy event frames that do not contain any structural information.
Due to these noisy event frames, the task transfer performance degenerates. 
Therefore, we mitigate this effect by introducing the proposed image gradient loss.

\subsection*{A1.6 Task Loss}
For the task loss $\mathcal{L}_{\text{task}}$, we use the standard cross-entropy loss and the loss defined in Yolov3~\cite{Redmon18arxiv} for classification and object detection, respectively. 
As illustrated with $\mathcal{L}_{\text{task}}$ in Fig.3 in the manuscript, the task loss is applied twice, once in the image domain and once in the translated event domain.

\subsection*{A1.7 Training Procedure}
The final loss is presented in Equation~\ref{sup_final_loss}.
As common in adversarial training, we train the generator and discriminator in separate steps.
The network parameters are updated alternatingly by minimizing the following losses, whereby we train the generator for one step after two discriminator training steps.
\begin{equation} \label{sup_final_loss}
\begin{split}
L_{Gen}  =&  \mathcal{L}_{\text{lat,gen.}} +\mathcal{L}_\text{recons.,gen.} +  \mathcal{L}_{\text{cycle}} \\
         & + 2\mathcal{L}_{\text{augm}} + \mathcal{L}_{grad} + \mathcal{L}_{task}\\
         & + \mathcal{L}_{\text{smooth}} \\
L_{Dis} =& \mathcal{L}_{\text{lat,disc.}} + \mathcal{L}_\text{recons.,disc.} 
\end{split}
\end{equation}

\vspace{7pt}
\section*{A2 Ablation}
We ablated our proposed design choices with multiple experiments for the task of classification on N-Caltech101~\cite{Orchard15fns} as well as for the task of object detection on MVSEC~\cite{Zhu18ral}.

\subsection*{A2.1 Classification}
The experiments conducted on N-Caltech101 underline the improved performance of our proposed transfer framework compared to standard \emph{Unsupervised Domain Adaptation (UDA)} methods, as reported in Table~\ref{tab:ncaltech101_abl}.
\\ \indent In a first experiment, we have adapted our network to a simple cycle GAN framework~\cite{Zhu17iccv}, in which we predict events and grayscale images directly from a shared feature space (Simple cycle).
The significantly lower performance of 0.577 compared to 0.848 of our final approach verifies the need of considering prior sensor knowledge and the feature space split into a shared embedding space and an event-specific space for motion information.
\\ \indent In a second experiment, we use our final network to solely translate from grayscale images to events (Ours transl).
This way, we generate a labeled event dataset on which a task network can be trained.
The performance of 0.832 verifies the accurate single image-to-event translation of our final network.
However, the lower performance compared to our final task transfer network confirms that the task network benefits from simultaneously learning on images and events to detect the most relevant task features.
A similar conclusion was also drawn in ~\cite{Sayed18gcpr}, where the authors achieved an increased task performance by learning with paired images and optical flow frames.
\\ \indent For the validation of our refinement network, we report the classification accuracy achieved without adding the residual event representation to the events constructed based on the flow and image gradients (Ours w/o ref).
As expected, the performance suffers a substantial drop compared to the final network, which shows the importance of the refinement network.
Due to the simple planar motion distribution, the event generation based on pseudo flow achieves the same performance as a direct prediction of the event representation (Ours w/o flow).
\\ \indent The benefits of the event generation based on the generative event model for more complex motions are shown in the experiments conducted on MVSEC.
Finally, as reported in the manuscript in Section M4.1, our transferred network achieves higher performance than the same architecture trained with ground truth labels.
This can be explained mainly for two reasons.
First, our UDA method allows us to include additional labeled image data to help train the event classification network.
Second, the simultaneous learning in the event and image representation increases the overall task performance.

\begin{table}
\centering
\begin{tabular}{m{3cm}C{1.2cm}>{\centering\arraybackslash}m{2cm}}

Method &  UDA & Accuracy $\uparrow$ \\
 \hline
Simple cycle & \greencheck & 0.577\\
Ours transl & \greencheck & 0.832 \\
Ours w/o ref & \greencheck & 0.592 \\
Ours w/o flow & \greencheck & \textbf{0.848} \\
Ours & \greencheck & \textbf{0.848} \\
\hline
Ours supervised & \redcross & 0.839 \\

\end{tabular}

\caption{\label{tab:ncaltech101_abl} Ablation for the classification task on the N-Caltech101 dataset.}
\end{table}

\begin{table}
\centering
\begin{tabular}{m{3cm}C{1.2cm}>{\centering\arraybackslash}m{2cm}}

Method & Unpaired & mAP $\uparrow$ \\
 \hline
Ours w/o flow & \greencheck & 0.26 \\
Ours w/o split & \greencheck & 0.41 \\
Ours w/o augm & \greencheck & 0.48 \\
Ours& \greencheck & \textbf{0.54} \\
\end{tabular}
\caption{\label{tab:mvsec_abl} Ablation for the object detecion task on the MVSEC dataset.}
\vspace{-15pt}
\end{table}

\subsection*{A2.2 Object Detection}
The ablation experiments for the task of objection detection on MVSEC strongly confirm our proposed network modules.
The first experiment (Ours w/o flow) shows that the image-to-event translation without the generative event model decreases the task transfer by a large margin.
The direct event representation prediction without the pseudo-flow estimation is not able to capture the complex motion distribution of a driving car sequence.
The second experiment (Ours w/o split) verifies the benefits of splitting the feature space into sensor shared and event-specific features. 
Without the embedding space split, the performance suffers a significant mAP drop from 0.54 to 0.41
In the third experiment (Ours w/o augm), we validate the introduced flow augmentations, which help to split the embedding space into shared and event-specific features.
The flow augmentations improve the detection score by 0.06 mAP.
In conclusion, each of our proposed network modules substantially improves the object detection performance.

%% file: root.bbl
\begin{thebibliography}{10}
\providecommand{\url}[1]{#1}
\csname url@rmstyle\endcsname
\providecommand{\newblock}{\relax}
\providecommand{\bibinfo}[2]{#2}
\providecommand\BIBentrySTDinterwordspacing{\spaceskip=0pt\relax}
\providecommand\BIBentryALTinterwordstretchfactor{4}
\providecommand\BIBentryALTinterwordspacing{\spaceskip=\fontdimen2\font plus
\BIBentryALTinterwordstretchfactor\fontdimen3\font minus
  \fontdimen4\font\relax}
\providecommand\BIBforeignlanguage[2]{{%
\expandafter\ifx\csname l@#1\endcsname\relax
\typeout{** WARNING: IEEEtran.bst: No hyphenation pattern has been}%
\typeout{** loaded for the language `#1'. Using the pattern for}%
\typeout{** the default language instead.}%
\else
\language=\csname l@#1\endcsname
\fi
#2}}

\bibitem{Fischer21online}
R.~Fisher, ``Cvonline: Image databases,''
  https://homepages.inf.ed.ac.uk/rbf/CVonline/Imagedbase.htm, 2021.

\bibitem{Yuhunang20eccv}
Y.~Hu, T.~Delbruck, and S.-C. Liu, ``Learning to exploit multiple vision
  modalitiesby using grafted networks,'' in \emph{Eur. Conf. Comput. Vis.
  (ECCV)}, 2020.

\bibitem{Gallego20pami}
G.~Gallego, T.~Delbruck, G.~Orchard, C.~Bartolozzi, B.~Taba, A.~Censi,
  S.~Leutenegger, A.~Davison, J.~Conradt, K.~Daniilidis, and D.~Scaramuzza,
  ``Event-based vision: A survey,'' \emph{{IEEE} Trans. Pattern Anal. Mach.
  Intell.}, 2020.

\bibitem{Rebecq19pami}
H.~Rebecq, R.~Ranftl, V.~Koltun, and D.~Scaramuzza, ``High speed and high
  dynamic range video with an event camera,'' \emph{{IEEE} Trans. Pattern Anal.
  Mach. Intell.}, 2019.

\bibitem{Zhu19arxiv}
A.~Z. Zhu, Z.~Wang, K.~Khant, and K.~Daniilidis, ``Eventgan: Leveraging large
  scale image datasets for event cameras,'' \emph{arXiv preprint
  arXiv:1912.01584}, 2019.

\bibitem{Gehrig20cvpr}
D.~Gehrig, M.~Gehrig, J.~Hidalgo-Carri\'o, and D.~Scaramuzza, ``{V}ideo to
  {E}vents: Recycling video datasets for event cameras,'' in \emph{{IEEE} Conf.
  Comput. Vis. Pattern Recog. (CVPR)}, 2020.

\bibitem{Hu21cvprw}
Y.~Hu, S.-C. Liu, and T.~Delbruck, ``{v2e} from video frames to realistic dvs
  events,'' in \emph{{IEEE} Conf. Comput. Vis. Pattern Recog. Workshops
  (CVPRW)}, 2021.

\bibitem{Russakovsky15ijcv}
O.~Russakovsky, J.~Deng, H.~Su, J.~Krause, S.~Satheesh, S.~Ma, Z.~Huang,
  A.~Karpathy, A.~Khosla, M.~Bernstein, A.~C. Berg, and F.-F. Li, ``{ImageNet}
  large scale visual recognition challenge,'' \emph{Int. J. Comput. Vis.}, vol.
  115, no.~3, Apr. 2015.

\bibitem{Ever10ijcv}
M.~Everingham, L.~Van~Gool, C.~Williams, J.~Winn, and A.~Zisserman, ``The
  pascal visual object classes (voc) challenge,'' \emph{International Journal
  of Computer Vision}, vol.~88, 06 2010.

\bibitem{sun2020scalability}
P.~Sun, H.~Kretzschmar, X.~Dotiwalla, A.~Chouard, V.~Patnaik, P.~Tsui, J.~Guo,
  Y.~Zhou, Y.~Chai, B.~Caine, \emph{et~al.}, ``Scalability in perception for
  autonomous driving: Waymo open dataset,'' in \emph{Proceedings of the
  IEEE/CVF Conference on Computer Vision and Pattern Recognition}, 2020.

\bibitem{Rebecq18corl}
H.~Rebecq, D.~Gehrig, and D.~Scaramuzza, ``{ESIM}: an open event camera
  simulator,'' in \emph{Conf. on Robotics Learning (CoRL)}, 2018.

\bibitem{Lee20ijcv}
H.-Y. Lee, H.-Y. Tseng, Q.~Mao, J.-B. Huang, Y.-D. Lu, M.~K. Singh, and M.-H.
  Yang, ``Drit++: Diverse image-to-image translation viadisentangled
  representations,'' \emph{International Journal of Computer Vision}, 2020.

\bibitem{Zheng20eccv}
Z.~Zheng, Y.~Wu, X.~Han, and J.~Shi, ``Forkgan: Seeing into the rainy night,''
  in \emph{The IEEE European Conference on Computer Vision (ECCV)}, August
  2020.

\bibitem{Zhu17iccv}
J.-Y. Zhu, T.~Park, P.~Isola, and A.~A. Efros, ``Unpaired image-to-image
  translation using cycle-consistent adversarial networks,'' in \emph{{IEEE}
  Conf. Comput. Vis. Pattern Recog. (CVPR)}, 2017.

\bibitem{Zhu18ral}
A.~Z. Zhu, D.~Thakur, T.~Ozaslan, B.~Pfrommer, V.~Kumar, and K.~Daniilidis,
  ``The multivehicle stereo event camera dataset: An event camera dataset for
  {3D} perception,'' \emph{{IEEE} Robot. Autom. Lett.}, vol.~3, no.~3, July
  2018.

\bibitem{Orchard15fns}
G.~Orchard, A.~Jayawant, G.~K. Cohen, and N.~Thakor, ``Converting static image
  datasets to spiking neuromorphic datasets using saccades,'' \emph{Front.
  Neurosci.}, vol.~9, 2015.

\bibitem{Wilson20tist}
\BIBentryALTinterwordspacing
G.~Wilson and D.~J. Cook, ``A survey of unsupervised deep domain adaptation,''
  \emph{ACM Trans. Intell. Syst. Technol.}, vol.~11, no.~5, July 2020.
  [Online]. Available: \url{https://doi.org/10.1145/3400066}
\BIBentrySTDinterwordspacing

\bibitem{Shen18aaai}
J.~Shen, Y.~Qu, W.~Zhang, and Y.~Yu, ``Wasserstein distance guided
  representation learning for domain adaptation,'' in \emph{in Proc. 32nd AAAI
  Conf. Artif. Intell., New Orleans, FL, USA, Feb. 2018 2-7, 2018}, S.~A.
  McIlraith and K.~Q. Weinberger, Eds.\hskip 1em plus 0.5em minus 0.4em\relax
  {AAAI} Press, 2018.

\bibitem{Ganin15icml}
Y.~Ganin and V.~Lempitsky, ``Unsupervised domain adaptation by
  backpropagation,'' in \emph{Proceedings of the 32nd International Conference
  on International Conference on Machine Learning - Volume 37}, ser. ICML'15,
  2015.

\bibitem{LiPei18bmvc}
P.~Li, X.~Liang, D.~Jia, and E.~P. Xing, ``Semantic-aware grad-gan
  forvirtual-to-real urban scene adaption,'' in \emph{British Mach. Vis. Conf.
  (BMVC)}, 2018.

\bibitem{Shrivastava17cvpr}
A.~Shrivastava, T.~Pfister, O.~Tuzel, J.~Susskind, W.~Wang, and R.~Webb,
  ``Learning from simulated and unsupervised images through adversarial
  training,'' in \emph{{IEEE} Conf. Comput. Vis. Pattern Recog. (CVPR)}, 2017.

\bibitem{Ryuhei20eccv}
R.~Takahashi, A.~Hashimoto, M.~Sonogashira, and M.~Iiyama, ``Partially-shared
  variational auto-encoders for unsupervised domain adaptation with target
  shift,'' in \emph{Computer Vision -- ECCV 2020}.\hskip 1em plus 0.5em minus
  0.4em\relax Springer International Publishing, 2020.

\bibitem{Hong18cvpr}
W.~{Hong}, Z.~{Wang}, M.~{Yang}, and J.~{Yuan}, ``Conditional generative
  adversarial network for structured domain adaptation,'' in \emph{{IEEE} Conf.
  Comput. Vis. Pattern Recog. (CVPR)}, 2018.

\bibitem{Zanardi19icra}
A.~Zanardi, J.~Zilly, A.~Aumiller, A.~Censi, and E.~Frazzoli, ``Wormhole
  learning,'' in \emph{{IEEE} Int. Conf. Robot. Autom. (ICRA)}, 2019.

\bibitem{Yang19miccai}
J.~Yang, N.~C. Dvornek, F.~Zhang, J.~Chapiro, M.~Lin, and J.~S. Duncan,
  ``Unsupervised domain adaptation via disentangled representations:
  Application to cross-modality liver segmentation,'' in \emph{Medical Image
  Computing and Computer Assisted Intervention -- MICCAI 2019}, 2019.

\bibitem{Mostafavi19cvpr}
S.~Mostafavi~I., L.~Wang, Y.-S. Ho, and K.-J.~Y. Yoon, ``Event-based high
  dynamic range image and very high frame rate video generation using
  conditional generative adversarial networks,'' in \emph{{IEEE} Conf. Comput.
  Vis. Pattern Recog. (CVPR)}, 2019.

\bibitem{Mostafavi20cvpr}
S.~M. Mostafavi~I., J.~Choi, and K.-J. Yoon, ``Learning to super resolve
  intensity images from events,'' in \emph{{IEEE} Conf. Comput. Vis. Pattern
  Recog. (CVPR)}, June 2020.

\bibitem{Wang20cvpr}
L.~Wang, T.-K. Kim, and K.-J. Yoon, ``Eventsr: From asynchronous events to
  image reconstruction, restoration, and super-resolution via end-to-end
  adversarial learning,'' in \emph{{IEEE} Conf. Comput. Vis. Pattern Recog.
  (CVPR)}, 2020.

\bibitem{Zhang20eccv}
S.~Zhang, Y.~Zhang, Z.~Jiang, D.~Zou, J.~Ren, and B.~Zhou, ``Learning to see in
  the dark with events,'' in \emph{Eur. Conf. Comput. Vis. (ECCV)}, 2020.

\bibitem{Wang21cvpr}
L.~Wang, Y.~Chae, S.-H. Yoon, T.-K. Kim, and K.-J. Yoon, ``Evdistill:
  Asynchronous events to end-task learning via bidirectional
  reconstruction-guided cross-modal knowledge distillation,'' in \emph{{IEEE}
  Conf. Comput. Vis. Pattern Recog. (CVPR)}, 2021.

\bibitem{Gallego15arxiv}
G.~Gallego, C.~Forster, E.~Mueggler, and D.~Scaramuzza, ``Event-based camera
  pose tracking using a generative event model,'' 2015, arXiv:1510.01972.

\bibitem{Huang18eccv}
X.~Huang, M.-Y. Liu, S.~Belongie, and J.~Kautz, ``Multimodal unsupervised
  image-to-image translation,'' in \emph{ECCV}, 2018.

\bibitem{Chen19cvpr}
Y.-C. Chen, Y.-Y. Lin, M.-H. Yang, and J.-B. Huang, ``Crdoco: Pixel-level
  domain transfer with cross-domain consistency,'' in \emph{{IEEE} Conf.
  Comput. Vis. Pattern Recog. (CVPR)}, 2019.

\bibitem{Hoffman17icml}
J.~Hoffman, E.~Tzeng, T.~Park, J.-Y. Zhu, P.~Isola, K.~Saenko, A.~A. Efros, and
  T.~Darrell, ``Cycada: Cycle consistent adversarial domain adaptation,'' in
  \emph{International Conference on Machine Learning (ICML)}, 2018.

\bibitem{Murez18cvpr}
Z.~{Murez}, S.~{Kolouri}, D.~{Kriegman}, R.~{Ramamoorthi}, and K.~{Kim},
  ``Image to image translation for domain adaptation,'' in \emph{2018 IEEE/CVF
  Conference on Computer Vision and Pattern Recognition}, 2018.

\bibitem{Rebecq19cvpr}
H.~Rebecq, R.~Ranftl, V.~Koltun, and D.~Scaramuzza, ``Events-to-video: Bringing
  modern computer vision to event cameras,'' in \emph{{IEEE} Conf. Comput. Vis.
  Pattern Recog. (CVPR)}, 2019.

\bibitem{Goodfellow14nips}
I.~Goodfellow, J.~Pouget-Abadie, M.~Mirza, B.~Xu, D.~Warde-Farley, S.~Ozair,
  A.~Courville, and Y.~Bengio, ``Generative adversarial nets,'' in \emph{Conf.
  Neural Inf. Process. Syst. (NIPS)}, 2014.

\bibitem{Isola17cvpr}
P.~Isola, J.-Y. Zhu, T.~Zhou, and A.~A. Efros, ``Image-to-image translation
  with conditional adversarial networks,'' in \emph{{IEEE} Conf. Comput. Vis.
  Pattern Recog. (CVPR)}, 2017.

\bibitem{Tran17nips}
D.~Tran, R.~Ranganath, and D.~M. Blei, ``Hierarchical implicit models and
  likelihood-free variational inference,'' in \emph{Proceedings of the 31st
  International Conference on Neural Information Processing Systems}, ser.
  NIPS'17.\hskip 1em plus 0.5em minus 0.4em\relax Red Hook, NY, USA: Curran
  Associates Inc., 2017.

\bibitem{Li06pami}
L.~Fei-Fei, R.~Fergus, and P.~Perona, ``One-shot learning of object
  categories,'' \emph{{IEEE} Trans. Pattern Anal. Mach. Intell.}, vol.~28,
  no.~4, 2006.

\bibitem{He16cvpr}
K.~He, X.~Zhang, S.~Ren, and J.~Sun, ``Deep residual learning for image
  recognition,'' in \emph{{IEEE} Conf. Comput. Vis. Pattern Recog. (CVPR)},
  2016.

\bibitem{Maqueda18cvpr}
A.~I. Maqueda, A.~Loquercio, G.~Gallego, N.~Garc\'ia, and D.~Scaramuzza,
  ``Event-based vision meets deep learning on steering prediction for
  self-driving cars,'' in \emph{{IEEE} Conf. Comput. Vis. Pattern Recog.
  (CVPR)}, 2018.

\bibitem{Sayed18gcpr}
\BIBentryALTinterwordspacing
N.~Sayed, B.~Brattoli, and B.~Ommer, ``Cross and learn: Cross-modal
  self-supervision,'' in \emph{German Conference on Pattern Recognition (GCPR)
  (Oral)}, Stuttgart, Germany, 2018. [Online]. Available:
  \url{https://arxiv.org/abs/1811.03879v1}
\BIBentrySTDinterwordspacing

\bibitem{Sironi18cvpr}
A.~Sironi, M.~Brambilla, N.~Bourdis, X.~Lagorce, and R.~Benosman, ``{HATS}:
  Histograms of averaged time surfaces for robust event-based object
  classification,'' in \emph{{IEEE} Conf. Comput. Vis. Pattern Recog. (CVPR)},
  2018.

\bibitem{Gehrig19iccv}
D.~Gehrig, A.~Loquercio, K.~G. Derpanis, and D.~Scaramuzza, ``End-to-end
  learning of representations for asynchronous event-based data,'' in
  \emph{Int. Conf. Comput. Vis. (ICCV)}, 2019.

\bibitem{Binas17icml}
J.~Binas, D.~Neil, S.-C. Liu, and T.~Delbruck, ``{DDD}17: End-to-end {DAVIS}
  driving dataset,'' in \emph{{ICML} Workshop on Machine Learning for
  Autonomous Vehicles}, 2017.

\bibitem{Brandli14ssc}
C.~Brandli, R.~Berner, M.~Yang, S.-C. Liu, and T.~Delbruck, ``A 240x180 130{dB}
  3$\mu$s latency global shutter spatiotemporal vision sensor,'' \emph{{IEEE}
  J. Solid-State Circuits}, vol.~49, no.~10, 2014.

\bibitem{Perot20nips}
E.~Perot, P.~de~Tournemire, D.~Nitti, J.~Masci, and A.~Sironi, ``Learning to
  detect objects with a 1 megapixel event camera,'' in \emph{Conf. Neural Inf.
  Process. Syst. (NIPS)}, 2020.

\bibitem{Redmon18arxiv}
\BIBentryALTinterwordspacing
J.~Redmon and A.~Farhadi, ``Yolov3: An incremental improvement,'' \emph{ar{X}iv
  e-prints}, vol. abs/1804.02767, 2018. [Online]. Available:
  \url{http://arxiv.org/abs/1804.02767}
\BIBentrySTDinterwordspacing

\bibitem{Zhu18rss}
A.~Z. Zhu, L.~Yuan, K.~Chaney, and K.~Daniilidis, ``{EV-FlowNet}:
  Self-supervised optical flow estimation for event-based cameras,'' in
  \emph{Robotics: Science and Systems (RSS)}, 2018.

\bibitem{Nikolic14icra}
J.~Nikolic, J.~Rehder, M.~Burri, P.~Gohl, S.~Leutenegger, P.~Furgale, and
  R.~Siegwart, ``A synchronized visual-inertial sensor system with {FPGA}
  pre-processing for accurate real-time {SLAM},'' in \emph{{IEEE} Int. Conf.
  Robot. Autom. (ICRA)}, 2014.

\bibitem{Sun14ijcv}
D.~Sun, S.~Roth, and M.~J. Black, ``A quantitative analysis of current
  practices in optical flow estimation and the principles behind them,''
  \emph{Int. J. Comput. Vis.}, 2014.

\end{thebibliography}
